\documentclass{article}

\usepackage{arxiv}

\usepackage[utf8]{inputenc} 
\usepackage[T1]{fontenc}    
\usepackage{hyperref}       
\usepackage{url}            
\usepackage{amsfonts}       
\usepackage{nicefrac}       
\usepackage{microtype}      
\usepackage{multirow}
\usepackage{tablefootnote}
\usepackage{footnote}
\usepackage{graphicx}
\usepackage{afterpage}
\usepackage[numbers,sort&compress]{natbib}
\usepackage{fullpage}
\usepackage{caption}
\usepackage{needspace}
\usepackage{times}
\usepackage{fancyhdr,graphicx,amsmath,amssymb}
\usepackage[linesnumbered,ruled]{algorithm2e}

\title{Shapelets for Earthquake Detection}

\author{
  Monica Arul\\
  NatHaz Modeling Laboratory\\
  Department of Civil Engineering\\
  University of Notre Dame\\
  Notre Dame, IN 46556 \\
  \texttt{maruljay@nd.edu} \\
   \And
 Ahsan Kareem \\
  NatHaz Modeling Laboratory\\
  Department of Civil Engineering\\
 University of Notre Dame \\
  Notre Dame, IN 46556 \\
  \texttt{kareem@nd.edu} \\
}

\begin{document}
\maketitle

\begin{abstract}
This paper introduces EQShapelets (EarthQuake Shapelets) a time-series shape-based approach embedded in machine learning to autonomously detect earthquakes. It promises to overcome the challenges in the field of seismology related to automated detection and cataloging of earthquakes. EQShapelets are amplitude and phase-independent, i.e., their detection sensitivity is irrespective of the magnitude of the earthquake and the time of occurrence. They are also robust to noise and other spurious signals. The detection capability of EQShapelets is tested on one week of continuous seismic data provided by the Northern California Seismic Network (NCSN) obtained from a station in central California near the Calaveras Fault. EQShapelets combined with a Random Forest classifier, detected all of the cataloged earthquakes and 281 
uncataloged events with lower false detection rate thus offering a better performance than autocorrelation and FAST algorithms. The primary advantage of EQShapelets over competing methods is the interpretability and insight it offers. Shape-based approaches are intuitive, visually meaningful and offers immediate insight into the problem domain that goes beyond their use in accurate detection. EQShapelets, if implemented at a large scale, can significantly reduce catalog completeness magnitudes and can serve as an effective tool for near real-time earthquake monitoring and cataloging. 
\end{abstract}


\section{Introduction}
The seismological field has witnessed over the years, a tremendous increase in the volume of seismic data due to the rapidly growing seismic networks and stations that can continuously record ground-motion measurements. The amount of data archived at the IRIS-DMC (Incorporated Research Institutions for Seismology Data Management Center) has grown by five folds in the last decade. As of July 2019, the IRIS-DMC has nearly 500 TB of seismic data. Given the large volume of continuous data, the automated detection of earthquakes is a longstanding problem for seismologists with the first attempt made in 1978 \cite{allen1978automatic}. Among the various methods developed in the past few decades to detect earthquakes, Short-term average/Long-term average (STA/LTA) \cite{allen1978automatic,withers1998comparison} and similarity-based search \cite{gibbons2006detection,shelly2007non,peng2009migration,kato2012propagation,skoumal2014optimizing,ross2017aftershocks,beauce2017fast,chamberlain2018eqcorrscan} are the most commonly used algorithms. 

While the traditional approaches using STA/LTA algorithms are generalized and efficient, they are highly sensitive to noise and fail to detect weak earthquakes and overlapping events. The similarity-based search methods use known waveforms as templates to search through the seismic data to find new events. Although the detection accuracy of these methods is higher, the need for prior knowledge of templates and the several correlation procedures involved render these methods computationally extensive and infeasible for large datasets. In recent years, there has been growing interest in using Machine Learning (ML) for automated earthquake detection and picking of seismic arrivals from earthquake data (e.g., \cite{dai1995automatic,wang1995artificial,tiira1999detecting,zhao1999artificial,wiszniowski2014application,perol2018convolutional,ross2018generalized,wu2018deepdetect}) Recently, a data-mining based approach known as FAST (Fingerprinting and Similarity Thresholding) \cite{yoon2015earthquake,bergen2018detecting} uses key discriminative features of waveforms to group and extract earthquake events. Although this algorithm is efficient, the detection results are limited to repeating earthquake events.

To overcome these limitations, we propose here a time-series shape-based approach for earthquake detection named EQShapelets (EarthQuake Shapelets).  The EQShapelets not only provide insights into the local patterns in the seismic time-series data but also can be used as input to ML classifiers to detect earthquakes from continuous time-histories and attain detection accuracies comparable to the state-of-the-art approaches. Unlike other similarity-based approaches which are restricted to detecting just the replicas of previously recorded events, EQShapelets can also detect unknown earthquake events. The concept of shapelets was first presented by \cite{ye2009time} as phase-independent time-series sub-sequences that are highly discriminative in predicting the target variable. Ever since the introduction of shapelets, researchers have used shapelets for time-series mining tasks such as characterization, clustering, classification and anomaly detection \cite{mueen2011logical,ye2011time,lines2012shapelet,zakaria2012clustering,grabocka2014learning,hills2014classification,beggel2019time}. In this paper, we adopt the shapelets algorithm to (a) discover EQShapelets, the discriminative segments in seismic time-series and (b) use the discovered EQShapelets to detect earthquakes from continuous time-histories. The detection sensitivity of EQShapelets is tested on a continuous seismic data set provided by the Northern California Seismic Network (NCSN) obtained from a station in central California near the Calaveras Fault. The results are then compared with results obtained by \cite{yoon2015earthquake} using the FAST approach. The algorithm and methodology involved are explained in great detail in the subsequent sections.

\section{Overview of time series shapelets}

Shapelets are based on a local shape-based approach for analyzing and classifying time series that focuses on highly informative subsequences of time series. Consider time series generated as a result of two events A and B as shown in Fig. 1. Both the time series have long stretches of aperiodic waveforms. However, a local shape appears for a short duration that differs substantially from the rest of the time series. These localized shapes are called shapelets. These discriminatory shapes which are phase and amplitude-independent serve as a powerful feature for time series mining tasks.

\begin{figure}[h]
  \centering
  \captionsetup{justification=centering}
  \includegraphics{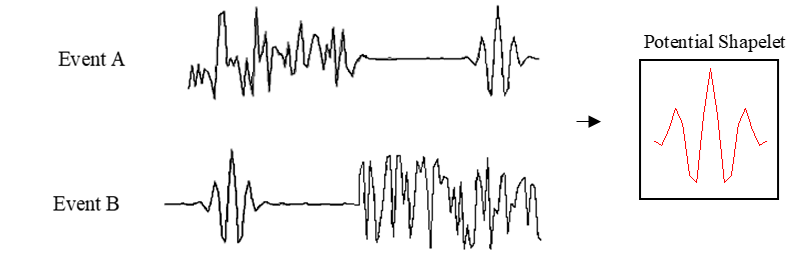}
  \caption{Time series shapelets}
  \label{fig:fig1}
\end{figure}

Analysis methods based on the global attributes of time series are unintuitive and reduce comprehensibility. By examining local-shape-based features, we ensure that these small discriminatory shapes are not averaged out but rather used to distinguish the time series, exactly as they are under intuitive visual inspection. Shapelets are not only helpful for time series classification but can also enhance understanding of time-series data for domain experts. 

Classification of time series based on shapelets uses a similarity measure between the time series and the shapelet, in the present case Euclidean distance, as a discriminatory feature to classify time series. Distance calculations (as outlined in the preliminaries section) decide the presence or absence of a shapelet in a particular time series. Each subsequence in each time series is considered as a potential shapelet candidate. Thus shapelets are found via an extensive search for every possible candidate of all possible lengths in every time series.

\section{Preliminaries}

\textbf{Definition 1: Time series dataset}:  A seismic time-series dataset $T=\left\{ {{T}_{1}},{{T}_{2,}}.....,{{T}_{n}} \right\}$ is a set of \textit{n} time series where a time series ${{T}_{i}}=\left\langle {{t}_{i,1}},{{t}_{i,2}},...,{{t}_{i,m}} \right\rangle $ is an ordered set of \textit{m} real numbers.

\textbf{Definition 2: Time series subsequence}: A length \textit{l} subsequence of ${{T}_{i}}$ is an ordered set of \textit{l} adjacent values from ${{T}_{i}}$. ${{T}_{i}}$ has a set ${{W}_{i,l}}$ of $\left( m-l \right)+1$ subsequences of length \textit{l}. Each subsequence $w=\left\{ {{t}_{j}},{{t}_{j+1}},...,{{t}_{j+l}} \right\}$ in ${{W}_{i,l}}$ is a time series of length \textit{l} where $1\le j<m-l$.

\textbf{Definition 3: Distance between subsequences}: The squared Euclidean distance between a subsequence X of length l and another subsequence Y of the same length l is defined as:
\begin{equation}
	\ dist(X,Y)=\sum\limits_{i=1}^{l}{{{\left( {{x}_{i}}-{{y}_{i}} \right)}^{2}}}\
\end{equation}

\textbf{Definition 4: EQShapelets}: An EQShapelet, \textit{S} is a subsequence of a seismic time series \textit{T}, that is discriminative of the class of the series.

\textbf{Definition 5: Learning set}: For a seismic time-series dataset \textit{T} and a set of corresponding class labels \textit{C} of the same size, a time series learning set $\Phi \left\{ T,C \right\}$ is given by a vector of instance input-output pairs ${{\Phi }_{i}}=({{T}_{i}},{{C}_{i}})$. 

\textbf{Definition 6: Time series classification}: Time series classification is the task of learning a classification function such that the predicted class labels (Ĉ) are closer to the original time series class labels (C).

\section{Dataset}

We selected the same set of data and adopted the exact same preprocessing techniques used in \cite{yoon2015earthquake} to detect earthquakes. One week of continuous earthquake waveform data (8th January 2011,00:00:00 to 15th January 2011,00:00:00) measured near the Calaveras Fault by the NC network at station CCOB.EHN is extracted from the Northern California Seismic Network (NCSN). The selected week of continuous data from the Northern California Earthquake Data Center (NCEDC) contains discontinuities in time series records. A total of 7 time gaps is noted with 14 minutes being the longest break in the time history record. The time series data is stitched together by omitting the missing time series records. The continuous data is then preprocessed by applying a 4- to 10-Hz bandpass filter to remove noise at lower frequencies. The denoised dataset is decimated from its original sampling frequency of 100 Hz to 20 Hz. 
It is very important to select the right time window to segment the continuous time-histories for mainly two reasons. A brute search for shapelets in n time series of length m has a complexity of $O({{n}^{2}}{{m}^{4}})$. Hence a large time window will render the method untenable. Secondly, the dataset under consideration is ridden with low-amplitude noise that can be easily mistaken for weak earthquake events. So a small time window of 20 s as used in \cite{yoon2015earthquake} will not provide seismically relevant shapelets as the method is based on capturing any distinct rise or fall in the time-series waveform in the given window. Apart from continuous time histories, NCSN also provides a catalog of 24 earthquake events and aftershocks that happened on the Calaveras Fault between 8 and 15 January 2011 with their time of occurrence, magnitude, and location. Upon close inspection of these cataloged events, a time window of 5 minutes was able to distinctly capture the occurrence of earthquake events and aftershocks in most cases. Hence a time interval of 5 minutes is chosen and the continuous seismic time series is broken down into 5-minute chunks. Thus each time history now contains 6000 data points and a total of 2004 time series datasets are obtained this way for the 1-week period under consideration. These 2004 time series records are used for the discovery of EQShapelets and detection of earthquakes in the following sections.

\section{Discovery of EQshapelets}

The discovery of EQShapelets has three major stages: generation of shapelet candidates, distance calculation between an EQshapelet and a seismic time series, and assessment of EQShapelet quality as shown in Fig. 2.

\begin{figure}[b]
  \centering
  \captionsetup{justification=centering}
  \includegraphics[scale=0.7]{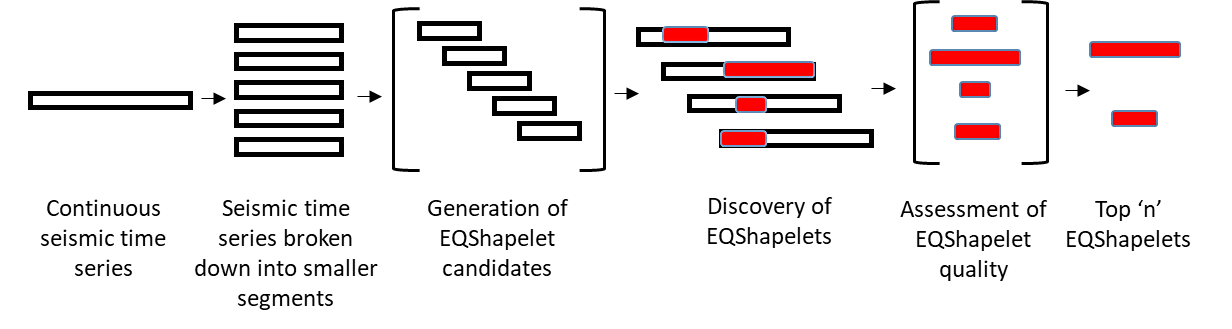}
  \caption{The process of discovery of EQShapelets}
  \label{fig:fig2}
\end{figure}

\subsection{Generation of EQShapelet candidates}
As shown in Fig.3, each subsequence in each time series in T is considered as a potential shapelet candidate. For a subsequence X of length l of a time series T of length m, the time series contains $(m-l)+1$ discrete subsequences of length l. If W is the set of all candidate shapelets of length l in a time series, then 

\begin{equation}
	\ W=\left\{ {{w}_{\min }},{{w}_{\min +1}},...,{{w}_{\max }} \right\}\
\end{equation}

where   as three is the minimum meaningful length and  . For example, in the present case, each seismic time series has 6000 data points. A minimum shapelet length of 3 generates 5998 candidate shapelets and a maximum shapelet length of 6000 generates 1 shapelet candidate. Thus the set of all candidate EQShapelets in a single seismic time series is given by

\begin{equation}
    \ {W}_{E{{Q}_{i}}}=\left\{ {{w}_{3}},{{w}_{4}},..,{{w}_{6000}} \right\}\
\end{equation}

\begin{figure}[h]
  \centering
  \captionsetup{justification=centering}
  \includegraphics[scale=0.8]{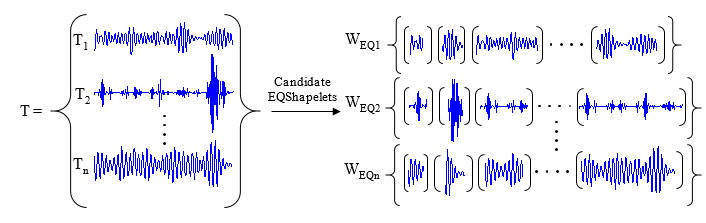}
  \caption{Illustration of generation of EQShapelet candidates for each time series in T}
  \label{fig:fig3}
\end{figure}

\subsection{Shapelet distance calculation}
The shapelet distance ${{d}_{S}}$ is the minimum squared Euclidean distance between an EQShapelet \textit{S} of length \textit{l} and any closest subsequence of length \textit{l} in \textit{T} i.e., distance between \textit{S} and its best matching location somewhere in \textit{T}, as shown in Fig.4. Thus the distance between a subsequence \textit{S}1 i.e., a potential EQShapelet candidate and all series in \textit{T} is computed to create a list of \textit{n} distances,

\begin{equation}
    	\ {{D}_{S}}=\left\langle {{d}_{S1,1}},{{d}_{S1,2}},...,{{d}_{S1,n}} \right\rangle \
\end{equation}

For clarity, this process is illustrated in Fig.5. It is a time-consuming task to calculate   and hence a number of speed-up techniques have been proposed in the literature to handle large volume of calculations \cite{ye2009time,mueen2011logical,rakthanmanon2013fast,hills2014classification}.

\subsection{Assessment of EQShapelet quality}
It is important to retain shapelet candidates that are seismically relevant and hence the shapelets are assessed for quality. Information Gain (IG) \cite{shannon1949themathematical} is usually used as the standard approach to calculate the quality of a shapelet \cite{ye2009time,mueen2011logical}. If a time series dataset \textit{TS} can be split into two classes, \textit{X} and \textit{Y} (ex: X = “Earthquake events” and Y = “Other”), then the entropy of \textit{TS} is: 

\begin{equation}
    \ H(TS)=-p(X)\log (p(X))-p(Y)\log (p(Y))\
\end{equation}
	
where \textit{p(X)} and \textit{p(Y)} are the proportion of time series objects in class \textit{X} and \textit{Y} respectively. Thus every splitting strategy partitions the dataset \textit{TS} into two sub-datasets ${{T}_{X}}$ and ${{T}_{Y}}$. The Information Gain of this split is the difference between the entropy of the entire dataset, and the sum of the weighted average of entropies for each split. As shown in Fig.6., in the present case, the distance to the EQShapelet is used as the splitting rule. Given a set of seismic time series dataset \textit{T}, an EQShapelet \textit{S} and a distance threshold ${{\overline{d}}_{t}}$, \textit{T} is split into two sub-datasets ${{T}_{a}}$ and ${{T}_{b}}$ such that for every time series in ${{T}_{a}}$, $dist({{T}_{a,i}},S)<{{\overline{d}}_{t}}$ and for every series in

\begin{figure}[p]
  \centering
  \captionsetup{justification=centering}
  \includegraphics[scale=0.85]{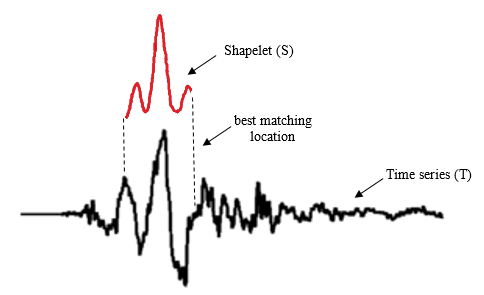}
  \caption{Illustration of best matching location for shapelet S in time series T}
  \label{fig:fig4}
\end{figure}

\begin{figure}[p]
  \centering
  \captionsetup{justification=centering}
  \includegraphics[scale=0.9]{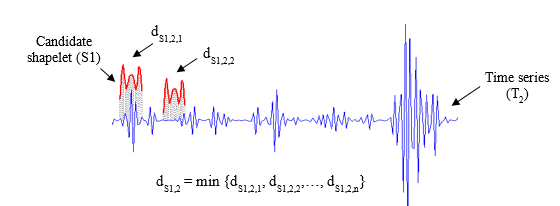}
  \caption{Illustration of Euclidean distance calculation between a candidate shapelet S1 and a time series T2}
  \label{fig:fig5}
\end{figure}

\begin{figure}[p]
  \centering
  \captionsetup{justification=centering}
  \includegraphics[scale=0.9]{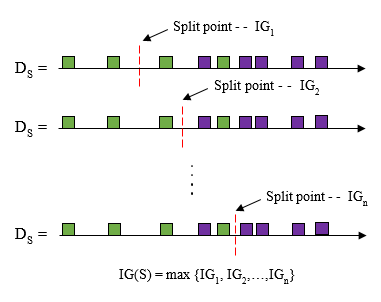}
  \caption{One-dimensional representation of the arrangement of time series objects by the distance to the candidate shapelet. Information Gain is calculated for each possible split point}
  \label{fig:fig6}
\end{figure}

${{T}_{b}}$, $dist({{T}_{b,i}},S)\ge {{\overline{d}}_{t}}$. Thus the information gain at each split point is calculated as:

\begin{equation}
    \ IG=H(T)-\left( \frac{|{{T}_{a}}|}{|T|}H({{T}_{a}})+\frac{|{{T}_{b}}|}{|T|}H({{T}_{b}}) \right)\
\end{equation}

where $0\le IG\le 1$.
The IG of an EQShapelet \textit{S}, is the highest IG of any split point and the EQShapelet with the highest IG has the most discriminative power. The IG calculation requires sorting the set of distances ${{D}_{S}}$ in Eq.(4) and then evaluating all possible split points as shown in Fig.6.

\subsection{Finding the EQShapelets}
An algorithm combining all of the above mentioned components of shapelet discovery was developed by \cite{bagnall2017great} and is available in GitHub (see Data and Resources Section). The same algorithm has been adopted and modified to suit the present case of discovery of EQShapelets. Algorithm 1 gives a pseudo-code overview of the process.The parameter \textit{n} in algorithm 1 represents the maximum number of EQShapelets to be stored and used for the detection of earthquake events. The value \textit{n} takes can have a significant impact on the run time and earthquake detection results. To find the optimal number of shapelets \textit{n}, an experiment is performed by varying quality i.e. the Information Gain (IG) threshold of the shapelets. This is carried out by segmenting a part of the seismic data into training and test set. A range of \textit{p} IG thresholds are used for the discovery of shapelets from the training set.  Thus \textit{p} different sets of \textit{n} shapelets are produced. A classifier is then trained using the \textit{p} sets of \textit{n} shapelets and detections are made on the test data. For each of the \textit{p} thresholds, \textit{p} classification accuracies and \textit{p} run time estimates, each corresponding to \textit{n}, the number of shapelets are obtained. The IG threshold and the corresponding \textit{n} shapelets with the best overall accuracy along with a low run time is selected for the discovery of EQShapelets.

\begin{algorithm}
\DontPrintSemicolon
\SetKwData{Left}{left}\SetKwData{This}{this}\SetKwData{Up}{up}
\SetKwFunction{Union}{Union}\SetKwFunction{FindCompress}{FindCompress}
\SetKwInOut{Input}{input}\SetKwInOut{Output}{output}
\Input{T (seismic time series) of length m, min (minimum EQShapelet length), max (maximum EQShapelet length), n (maximum number of EQShapelets to store, quality (predefined information gain threshold)}
\Output{EQShapelets}
\BlankLine
$nShapelets \longleftarrow \Phi$\;
$C \longleftarrow class$ $labels$ $(T)$\;
\ForAll{$T_{i}\in T$}{
$shapelets \longleftarrow \Phi$\;
\For{$l \longleftarrow min$ $to$ $max$}{
$W_{i,l} \longleftarrow generate$ $shapelet$ $candidates$ $(T_{i},min,max)$\;
\ForAll{$subsequences S\in  W_{i,l}$}{
$D_{S} \longleftarrow calculate$ $distances$ $(S, W_{i,l})$\;
$quality \longleftarrow evaluate$ $candidate$ $shapelets$ $(S, D_{S})$\;
\ $shapelets.add$ $(S, quality)$
}
}
\ $group$ $by$ $quality$ $(shapelets)$\;
\ $remove$ $similar$ $(shapelets)$\;
\ $EQShapelets \longleftarrow merge$ $(n, nShapelets, shapelets)$\;
}
\caption {Discovery of Shapelets}
\end{algorithm}

\section{Results}

\subsection{Training/testing sets}
According to the NCSN catalog, an Mw 4.1 earthquake occurred on the Calaveras Fault on 8 January 2011, followed by several aftershocks. Thus the continuous time history recorded between 8 January 2011 (00:00:00) and 9 January 2011 (00:00:00) is used for the discovery of EQShapelets.  As mentioned earlier, a time window of 5 minutes is used to segment the continuous time history into 288 smaller datasets. Out of the 288 datasets, 52 time histories corresponding to earthquake events are manually labeled as “Earthquake events”. 52 other noisy time-history records that do not contain any earthquake events are selected and labeled as “Other. These 104 labeled set of time series T serves as the ‘time-series learning set’ as mentioned in the preliminaries section. It can be noted that the learning set contains equal samples of “Earthquake events” and “other” This is done to achieve a balanced training set to avoid classifier bias during the detection of earthquakes. The dataset in then randomly split into training (60\%) and test (40\%) sets.

\subsection{EQShapelets}
The IG threshold experiment is performed for a range of 10 different IGs $(0.05,0.10,0.15,.....,0.50)$for the discovery of EQShapelets from the training data. A Random Forest classifier (with a constant classification threshold of 0.5) is trained based on the discovered shapelets and is used to make detections on the test set. The variation of the number of discovered EQShapelets, detection accuracy and run time with respect to IG threshold is shown in Fig. 7. 

\begin{figure}[h]
  \centering
  \captionsetup{justification=centering}
  \includegraphics[scale=0.6]{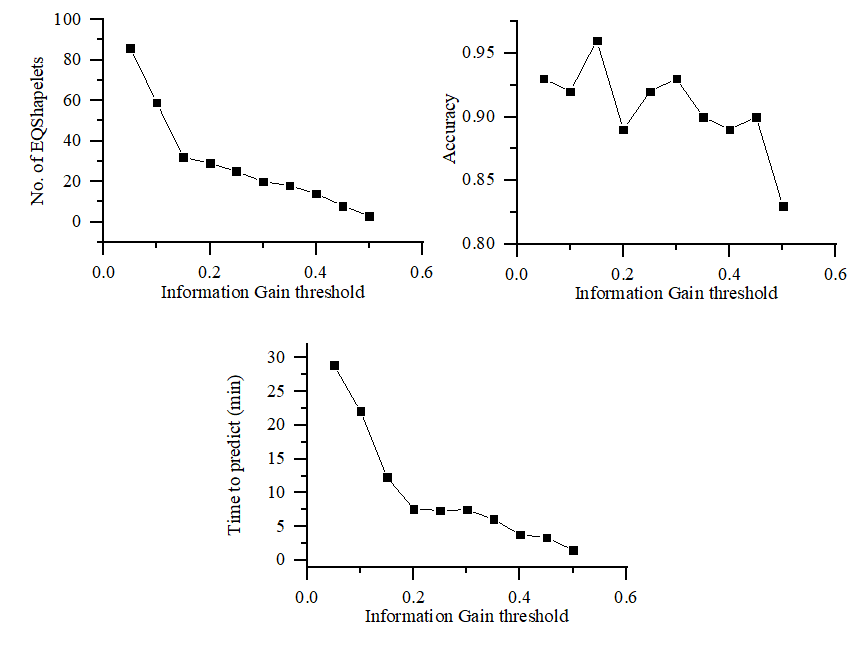}
  \caption{Effect of Information Gain threshold on the number of EQShapelets discovered, accuracy and run time for the detection of earthquake events}
  \label{fig:fig7}
\end{figure}

It can be seen that as the IG threshold increases, the number of discovered shapelets and the run time drastically decreases. The accuracy on the other hand keeps fluctuating. This is because, using too few shapelets would not provide the algorithm enough insights to make correct detections while using too many over fits the classifier making it to perform poorly on the test set. After careful inspection, an IG threshold of 0.45 is selected for the discovery of EQShapelets. This threshold produces 8 shapelets and obtains an accuracy of 90\% and a detection time of 3.4 minutes. Any threshold higher that this reduces the accuracy while a lower threshold results in the discovery of large number of shapelets thus drastically increasing the run time. 
The algorithm 1 with input parameters T, min = 3, max = 6000, n = 8 and quality = 0.45 is implemented in Python as a single-core serial job on an Intel Xeon Processor E5-2620 (2.6-GHz CPU). The algorithm takes 270 minutes with a memory usage of 5.9 GB to search through every time history in the training set to produce EQShapelets. The shapelet algorithm returns 8 time-history datasets containing the most discriminatory shapelets. The top 8 EQShapelets along with their respective IGs are shown in Fig 8. This serves as an interpretable result in how earthquake detections can be made using EQShapelets. The highlighted section of the series is the EQShapelet and it occurs with the onset of an impulsive waveform and stops after the end of the event. The top 8 EQShapelets that are extracted are from the time series shown in Fig. 9 along with their length. 

\begin{figure}
  \centering
  \captionsetup{justification=centering}
  \includegraphics[scale=0.8]{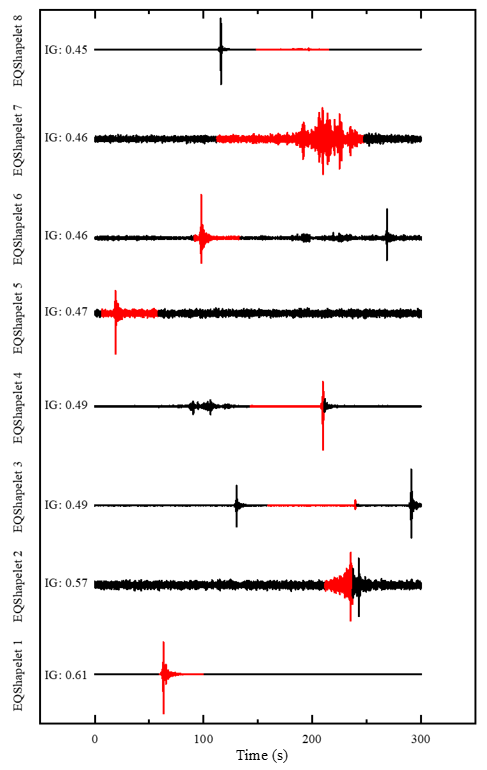}
  \caption{Discovery of EQShapelets from earthquake waveforms recorded on 8th January 2011 at station CCOB.EHN}
  \label{fig:fig8}
\end{figure}

\afterpage{
\begin{figure}
  \centering
  \captionsetup{justification=centering}
  \includegraphics[scale=0.9]{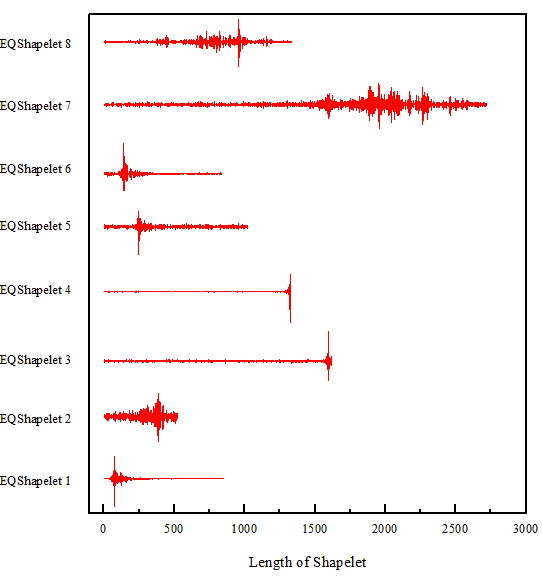}
  \caption{EQShapelets discovered from continuous data recorded on 8 January 2011, ordered by Information Gain (IG)}
  \label{fig:fig9}
\end{figure}
\clearpage
}

\section{Earthquake detection on continuous records using EQShapelets}

The detection of earthquakes from continuous time histories can be treated as a binary classification problem as to whether a time series contains earthquake event or not. The shapelet based classifier originally developed by \cite{ye2009time} embeds shapelet finding in a decision tree classifier where shapelets are found at every node. Many researchers ever since have demonstrated that higher accuracy can be achieved by using shapelets with more complex classifiers or ensemble of classifiers than with decision trees, where overfitting is a major issue \cite{lines2012shapelet,hills2014classification,bagnall2017great,bostrom2017binary}.
For the present case, Random Forest \cite{breiman2001random} is used as a classifier to detect earthquake events. The Random Forest algorithm seeks to solves the issues with decision trees by classifying examples through using multitude of decision trees and predicting the class of a sample based on the mean probability estimate across all the trees. Thus detection of earthquakes in continuous data has two stages. Firstly, the trained EQShapelet-based classifier searches for impulsive earthquake-like waveforms in every time series and then classifies a series as either containing earthquake event or not. Secondly, estimating the class probability for each prediction based on the mean predicted class probabilities of all the trees in the forest. For example, if a time series is predicted as A (A = “Earthquake events” and B = “Other”), the classifier also returns a probability for that prediction, i.e. prob(A) = 87\% and prob(B) = 13\%. This makes the detection of earthquakes more transparent and interpretable and helps the user make more informed decisions.

\subsection{Results and Discussion}
The EQShapelet-based Random Forest classifier is used to detect earthquake events from time history recorded between 9 January 2011 (00:00:00) and 15 January 2011 (00:00:00). A summary of the detection performance of EQShapelets in terms of several metrics is provided in Table 1. EQShapelets were able to detect a whopping total of 299 earthquake events between 9 and 15 January 2011 within 3 hours 29 minutes and 33 seconds. A prediction probability was also returned by the classifier for each of these detections. Fig.10 shows the number of event detections and their respective prediction probabilities. Out of the 299 events, 95 events were detected with a high prediction probability between 96\% and 100\% while 41 events buried in noise were detected with a probability of 50\% – 59\%. Nearly 46\% of the detected events has a prediction probability of 90\% and higher. EQShapelets were able to detect all the 13 catalog events between

\begin{figure}[htbp]
  \centering
  \captionsetup{justification=centering}
  \includegraphics[scale=0.8]{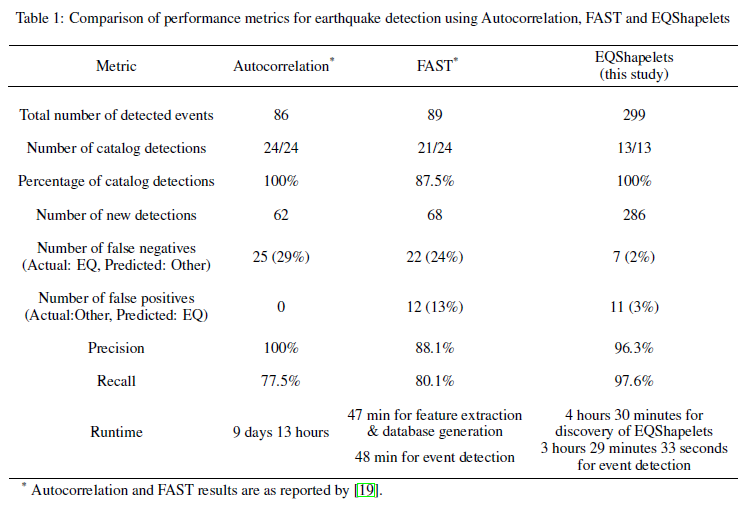}
\end{figure}

\begin{figure}[htbp]
  \centering
  \captionsetup{justification=centering}
  \includegraphics[scale=0.6]{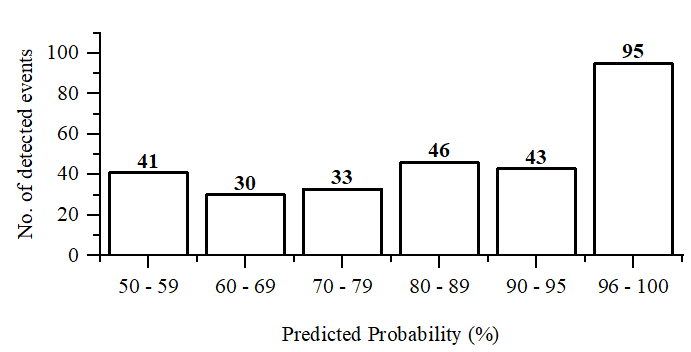}
  \caption{Number of event detections for various intervals of prediction probabilities}
  \label{fig:fig10}
\end{figure}

 9 and 15 January as shown in Fig.11. (Note: The NCSN catalog has 24 cataloged events out of which 11 events occurred on 8 January 2011 and rest of the 13 events occurred between 9 – 14 January 2011). In addition to the 13 catalog events, EQShapelet-based classifier was also able to detect 286 new events that were not available in the catalog. An example of new events detected with high and low prediction probabilities are shown in Fig.12 and Fig.13. respectively. It can be seen that the algorithm detects a large variety of events with different waveforms allowing it to generalizes well to waveforms that are not similar to the ones in the training set.

\subsection{Detection accuracy}
The new detected events are compared with signals on all three components (EHE, EHN, EHZ) of the data recorded at station CCOB as shown in Fig.14 – Fig. 17. The waveforms are carefully inspected to make sure that the waveforms classified as true earthquake events resemble an impulsive earthquake signal on all three channels although EQShapelets were only trained on the EHN channel for detection.

\begin{figure}[p]
  \centering
  \captionsetup{justification=centering}
  \includegraphics[scale=0.9]{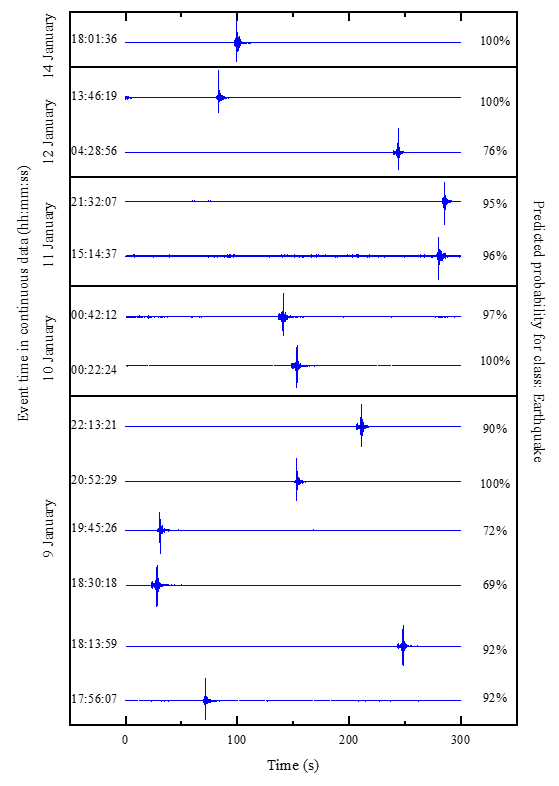}
  \caption{Cataloged earthquake waveforms detected by EQShapelets ordered by event time recorded between 9 and 14 January 2011 from CCOB.EHN}
  \label{fig:fig11}
\end{figure}

\begin{figure}[p]
  \centering
  \captionsetup{justification=centering}
  \includegraphics[scale=0.9]{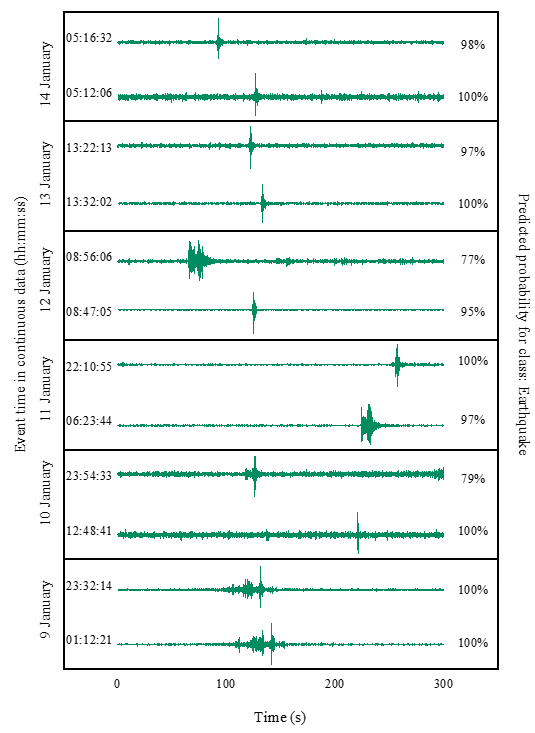}
  \caption{Example of waveforms predicted as earthquake events with high probability by EQShapelets ordered by event time recorded between 9 and 14 January 2011 from CCOB.EHN}
  \label{fig:fig12}
\end{figure}

\begin{figure}[p]
  \centering
  \captionsetup{justification=centering}
  \includegraphics[scale=0.9]{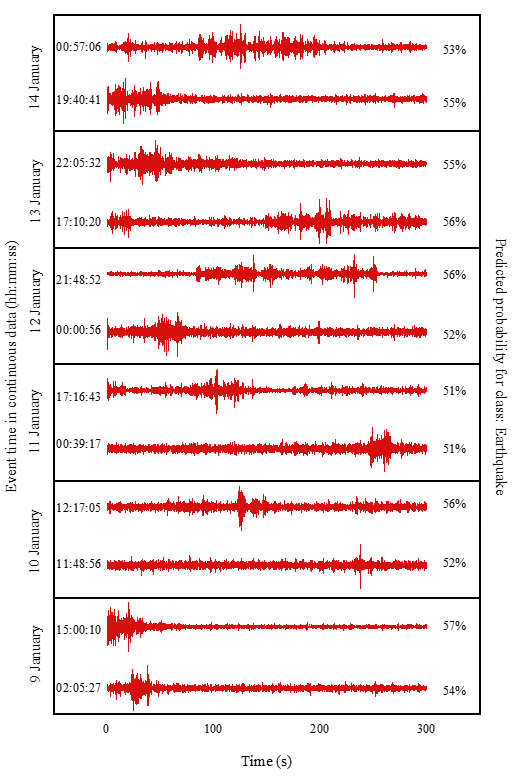}
  \caption{Example of waveforms predicted as earthquake events with low probability by EQShapelets ordered by event time recorded between 9 and 14 January 2011 from CCOB.EHN}
  \label{fig:fig13}
\end{figure}

\afterpage{
\begin{figure}
  \centering
  \captionsetup{justification=centering}
  \includegraphics[scale=0.9]{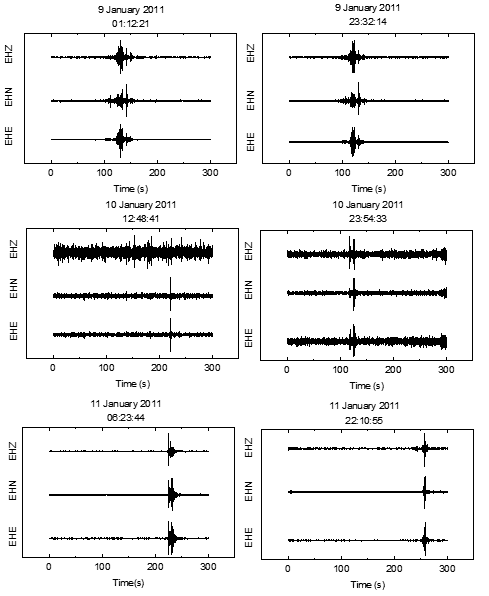}
  \caption{Event detections (9 – 11 January 2011) with high probability compared with the data from two other components (EHE, EHZ) at station CCOB to check if the detected events are truly impulsive earthquake waveforms}
  \label{fig:fig14}
\end{figure}
\clearpage
}

\afterpage{
\begin{figure}
  \centering
  \captionsetup{justification=centering}
  \includegraphics[scale=0.9]{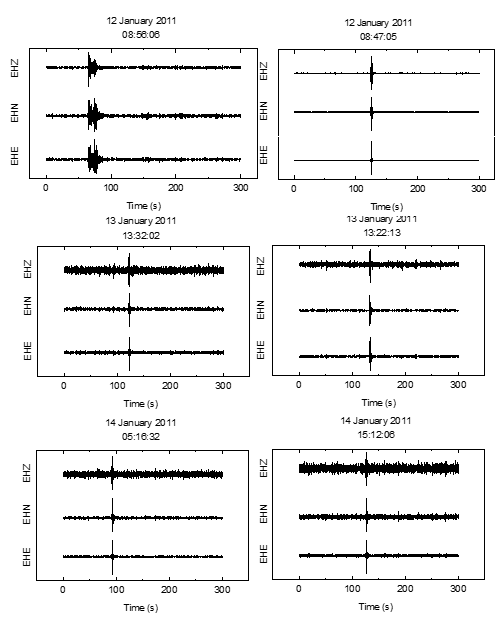}
  \caption{Event detections (12 – 14 January 2011) with high probability compared with the data from two other components (EHE, EHZ) at station CCOB to check if the detected events are truly impulsive earthquake waveforms}
  \label{fig:fig15}
\end{figure}
\clearpage
}

\afterpage{
\begin{figure}
  \centering
  \captionsetup{justification=centering}
  \includegraphics[scale=0.9]{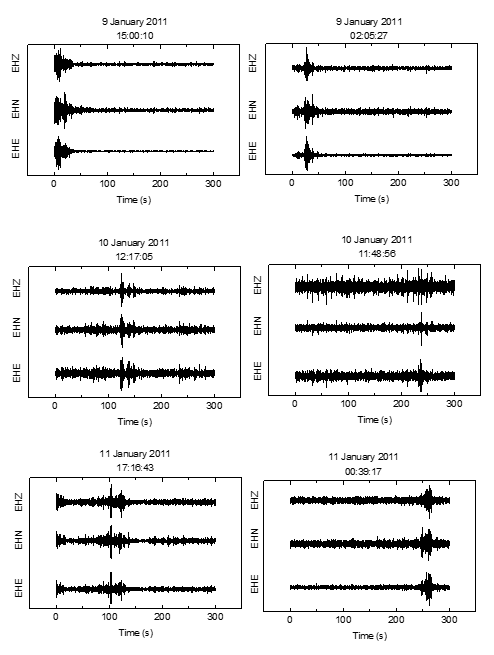}
  \caption{Event detections (9 – 11 January 2011) with low probability compared with the data from two other components (EHE, EHZ) at station CCOB to check if the detected events are truly impulsive earthquake waveforms.}
  \label{fig:fig16}
\end{figure}
\clearpage
}

\afterpage{
\begin{figure}
  \centering
  \captionsetup{justification=centering}
  \includegraphics[scale=0.9]{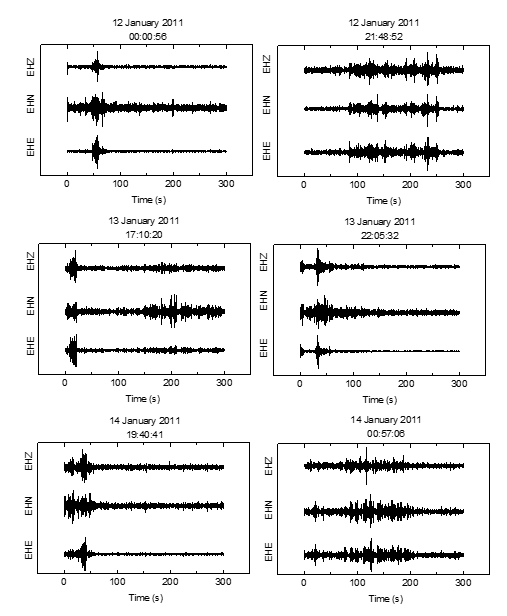}
  \caption{Event detections (12 – 14 January 2011) with low probability compared with the data from two other components (EHE, EHZ) at station CCOB to check if the detected events are truly impulsive earthquake waveforms.}
  \label{fig:fig17}
\end{figure}
\clearpage
}

The noisy events detected with a probability between 50\% to 60\% are carefully checked for the presence of false positives and false negatives. There were 7 false negatives (missed detections) and 11 false positives (wrong detections). Therefore, the proportion of detected events that are true events i.e., precision is 96.3\% and recall is 97.6\%. The detection sensitivity of EQShapelets is clearly high as the EQShapelet-based classifier which shows the robustness of EQShapelet-based classifier for identifying earthquake events.

\section{Comparison with other earthquake detection methods}

Table 1 reports the detection performance of autocorrelation, FAST and EQShapelets. All three methods use the same 1 week of continuous time series data obtained using single-channel seismogram (CCOB.EHN). EQShapelets were able to detect almost 200 more events with lower false negatives that the other two approaches. Also, all catalog events (13/13) were correctly detected by the EQShapelet-based classifier. In terms of false positives, EQShapelets returned 11 false positives which constitutes only 3\% of the total detections. On the other hand, autocorrelation and FAST had 29\% and 24\% false positive rate respectively. Thus the overall precision of EQShapelets is far better than FAST and comparable to autocorrelation.

In terms of run time, EQShapelets take significantly longer duration for training (i.e., discovery of shapelets), compared to FAST, due to the large volume of distance calculations involved. Unlike FAST, that extracts features and updates its database every time before event detection, the EQShapelet discovery is a one-time process. However, the time to discover EQShapelets can be further reduced by adopting one of the several speed-up techniques proposed in the literature \cite{ye2009time,ye2011time,mueen2011logical,rakthanmanon2013fast,hills2014classification}.

\section{Scalability to large datasets}

From Table, 1 it can be seen that EQShapelet-based classifier takes 3 hours 29 minutes 33 seconds
to detect earthquake events from 1 week of continuous data. The run time will drastically increase if event detections are made for months or years of continuous data. The run time for earthquake detection can be reduced by two ways. Incorporating parallelism in the algorithm so that distance calculations can be executed in parallel on multicore machines. Another way is to redesign the algorithm to make it suitable for parallel Graphics Process Units (GPUs). \cite{chang2012efficient} improved the shapelet algorithm for GPU implementation and achieved speedups nearly 2 orders of magnitude faster than CPU implementation. This means that a 1.7-hour CPU implementation of shapelets will only take 2 minutes using GPUs. Such an algorithm redesign to EQShapelets will render this method efficient for processing large volumes of seismic data.

\section{Conclusion and future implications}

Automated detection of earthquakes is a longstanding problem for seismologists. In this paper, we address this fundamental problem by autonomously identifying earthquake events in continuous time history using EQShapelets. One week of continuous earthquake waveform data (9 January 2011,00:00:00 to 15 January 2011,00:00:00) measured by the NC network at station CCOB.EHN near the Calaveras Fault is retrieved from the Northern California Seismic Network (NCSN). After preprocessing, the continuous time-history is broken down into 5-minute chunks resulting in a total of 2004 time series datasets. The continuous time series on 8 January 2011 is used to construct the learning set which yields the discovery of top 8 EQShapelets. The 8 EQShapelets are then used in combination with Random Forest classifier to detect earthquake events from continuous time histories between 9 and 15 January 2011. EQShapelets were able to detect 13/13 catalog events and also detected 286 new events in 3 hours 29 minutes and 33 seconds. This runtime can be greatly reduced in the future implementations by incorporating parallelism into the algorithm. EQShapelet-based classifier was able to achieve a precision of 96.3\% with a very low false positive and negative rates (3\% and 2\% respectively). On comparison with other detection methods such as autocorrelation and FAST, EQShapelets have a higher detection sensitivity and were able to detect 200 more events than the other two approaches.

The primary advantage of EQShapelets over competing methods is interpretability and insight. Shape-based approaches are intuitive, visually meaningful and offers immediate insight into the problem domain that goes beyond their use in accurate detection. Another important advantage of EQShapelets is that they are amplitude and phase-independent meaning that their detection sensitivity is irrespective of the magnitude of the earthquake and the time of occurrence in a given window. EQShapelets are perfectly capable of detecting weak earthquakes masked by noisy signals and unlike other similarity-based approaches which are limited to detecting only replicas of previously recorded events, EQShapelets can also detect unknown earthquake events. A shape-based approach such as EQShapelets can solve a wide range of research problems in seismology pertaining to shape of seismic waveforms, for example, detection and classification of seismic wave phases. EQShapelets, if implemented at a large scale, can significantly reduce catalog completeness magnitudes and can serve as an effective tool for near real-time earthquake monitoring and cataloging.

\section{Data and resources}

Continuous waveform data, and earthquake catalogs for this study were last accessed in August 2019 accessed through the Northern California Earthquake Data Center (NCEDC), doi:10.7932/NCEDC (Northern California Earthquake Data Center), operated by the UC Berkeley Seismological Laboratory and the U.S. Geological Survey (USGS). The algorithm for shapelet discovery is available at "Anthony Bagnall, Jason Lines, William Vickers and Eamonn Keogh, The UEA and UCR Time Series Classification Repository” (\url{www.timeseriesclassification.com}). Additional data related to this paper may be requested from the authors

\section*{Acknowledgement}
We thank the Northern California Earthquake Data Center for the earthquake catalog and continuous seismic waveform data and the University of Notre Dame’s Center for Research Computing for providing cluster computing resources. We would also like to thank Dr. Anthony Bagnall from the University of East Anglia for providing access to the shapelet codes.

\bibliographystyle{unsrt}  
\bibliography{references}  





\end{document}